\documentclass[sigconf]{acmart}

\usepackage{hyperref}
\usepackage{url}
\usepackage{appendix}
\usepackage{tocloft}

\usepackage{amsmath}
\usepackage{mathtools}
\usepackage{amsthm}
\usepackage{bbm}

\usepackage{tabularx}
\usepackage{booktabs}
\usepackage{makecell}
\usepackage{adjustbox}
\usepackage{multirow}
\usepackage{multicol}
\usepackage{colortbl}
\usepackage{wrapfig}
\usepackage{algorithm,algpseudocode}

\newcommand{\grayrow}{\rowcolor[gray]{.90}}

\AtBeginDocument{%
  }

\copyrightyear{2025}
\acmYear{2025}
\setcopyright{cc}
\setcctype{by}
\acmConference[KDD '25]{Proceedings of the 31st ACM SIGKDD Conference on Knowledge Discovery and Data Mining V.2}{August 3--7, 2025}{Toronto, ON, Canada}
\acmBooktitle{Proceedings of the 31st ACM SIGKDD Conference on Knowledge Discovery and Data Mining V.2 (KDD '25), August 3--7, 2025, Toronto, ON, Canada}
\acmDOI{10.1145/3711896.3737002}
\acmISBN{979-8-4007-1454-2/2025/08}

\begin{document}

\title{Improving Group Robustness on Spurious Correlation via Evidential Alignment}
\author{Wenqian Ye}
\affiliation{%
  \institution{University of Virginia}
  \city{Charlottesville}
  \state{VA}
  \country{USA}}
\email{wenqian@virginia.edu}

\author{Guangtao Zheng}
\affiliation{%
  \institution{University of Virginia}
  \city{Charlottesville}
  \state{VA}
  \country{USA}}
\email{gz5hp@virginia.edu}

\author{Aidong Zhang}
\affiliation{%
  \institution{University of Virginia}
  \city{Charlottesville}
  \state{VA}
  \country{USA}}
\email{aidong@virginia.edu}


\begin{abstract}

Deep neural networks often learn and rely on spurious correlations, i.e., superficial associations between non-causal features and the targets. For instance, an image classifier may identify camels based on the desert backgrounds. While it can yield high overall accuracy during training, it degrades generalization on more diverse scenarios where such correlations do not hold. This problem poses significant challenges for out-of-distribution robustness and trustworthiness. Existing methods typically mitigate this issue by using external group annotations or auxiliary deterministic models to learn unbiased representations. However, such information is costly to obtain, and deterministic models may fail to capture the full spectrum of biases learned by the models. To address these limitations, we propose Evidential Alignment, a novel framework that leverages uncertainty quantification to understand the behavior of the biased models without requiring group annotations. By quantifying the evidence of model prediction with second-order risk minimization and calibrating the biased models with the proposed evidential calibration technique, Evidential Alignment identifies and suppresses spurious correlations while preserving core features. We theoretically justify the effectiveness of our method as capable of learning the patterns of biased models and debiasing the model without requiring any spurious correlation annotations. Empirical results demonstrate that our method significantly improves group robustness across diverse architectures and data modalities, providing a scalable and principled solution to spurious correlations. 
\end{abstract}




\maketitle

\section{Introduction}

\begin{figure}[h]
    \centering
    \includegraphics[width=0.8\linewidth]{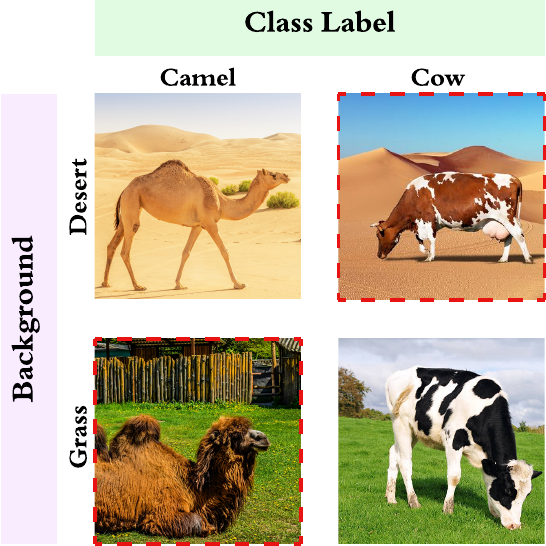}
    \caption{An example of cow/camel classification task. Red dotted lines denote the minority groups where spurious correlation does not hold.}
    \label{fig:example}
\end{figure}

Deep neural networks trained with empirical risk minimization (ERM)~\cite{vapnik1999overview} often exhibit spurious biases, i.e., the reliance on spurious correlations to make predictions. Spurious correlations are brittle associations between prediction targets and non-essential features (e.g., background, texture, or secondary objects)~\cite{stock2018convnets,baker2018deep,beery2018recognition,sagawadistributionally,ye_spurious_2024,zheng2024benchmarking}. These correlations can lead models to achieve high average accuracy but perform poorly on minority groups where the spurious correlations are absent. For example, consider a cow/camel classification task shown in Figure~\ref{fig:example}, the training set may be biased so that the camel class is spuriously correlated with desert backgrounds. 
ERM-trained models tend to rely on the spurious attributes--desert backgrounds--to identify the camel class. This may lead to biased predictions and poor generalization on minority groups where camels are on grass backgrounds~\cite{beery2018recognition}. The high discrepancy in generalization across groups caused by spurious correlations is especially problematic in high-stakes applications such as medicine~\cite{zech2018variable}, algorithmic fairness~\cite{khani2021removing}, and criminal justice~\cite{barocas2023fairness}, where biased predictions can have severe consequences.

Therefore, it is crucial to improve a model's \textit{group robustness}, i.e., ensuring generalization across data groups with varying spurious correlations. Group robustness is typically measured by the model's worst accuracy across groups~\cite{sagawadistributionally}. In practice, the groups are formulated with group labels that pair samples' class labels with their associated spurious attributes, explicitly indicating the correlations that a model may learn. The group labels can be used to construct balanced groups~\cite{kirichenko2022last,zheng_learning_2024,zheng_spuriousness-aware_2024} or minimize the worst-group loss~\cite{sagawadistributionally} to learn invariant representations. Moreover, studies have demonstrated that retraining classifiers on a group-balanced held-out dataset or on samples representing the failures of a biased model can significantly enhance group robustness~\cite{liu2021just,nam2020learning,yao2022improving,kim2022learning,zheng2025shortcutprobe}. However, obtaining group labels can be prohibitively expensive in many real-world applications and these labels sometimes cannot accurately reflect the subtle biases learned by a model. To address these challenges, recent research has focused on group label estimation with biased auxiliary models or sample reweighting mechanisms as alternative strategies for improving a model’s group robustness.

Despite significant progress in improving group robustness, existing methods typically rely on external group annotations, which are both costly and labor-intensive to obtain. In addition, most existing annotation-free approaches use deterministic procedures to identify and mitigate spurious biases. This offers only a narrow perspective on how models exploit spurious correlations. Consequently, these methods may not capture the full range of biases present in real-world data. These limitations underscore the need for scalable, annotation-free approaches that provide a more comprehensive understanding of spurious biases.

To overcome the limitations of current group robustness techniques, we propose a novel framework equipped with uncertainty estimation to gain deeper insights into spurious biases. Uncertainty can be commonly divided into aleatoric (data) and epistemic (model) uncertainty.  In particular, we focus on the epistemic (model) uncertainty as it provides critical information about a model’s confidence when it encounters ambiguous data~\cite{kendall2017uncertainties,hullermeier2021aleatoric}. Such information provides insights into the data in minority groups, which is challenging for a model with spurious biases and is the bottleneck in the model's group robustness. We leverage the Dempster–Shafer Theory of Evidence (DST)~\cite{Yager2010ClassicWO} to provide a general and well-defined theoretical framework for reasoning with uncertainty. Recent advances in uncertainty quantification~\cite{sensoy2018evidential,amini2020deep} have introduced scalable methods for estimating model uncertainty in a single-pass nature without requiring explicit priors such as group annotations. Building on these insights, we propose \textbf{Evidential Alignment}, a novel framework that enhances group robustness without group annotations.

Our approach first augments traditional logit-based prediction probabilities with an uncertainty measure by transforming the objective of a biased ERM model from \textit{first-order risk minimization} to \textit{second-order risk minimization}, which refers to learning a distribution over distributions. By modeling uncertainty with a Dirichlet distribution, the model produces both prediction probabilities and an uncertainty estimate. We then apply an evidential calibration procedure to suppress spurious biases while preserving core features via an automatic reweighting process with a calibration dataset. Unlike traditional approaches that depend on external annotations or deterministic bias estimation, Evidential Alignment leverages model uncertainty to automatically detect and mitigate spurious correlations. This uncertainty-guided strategy provides a data-driven measure to pinpoint where the model excessively relies on non-causal features, enabling dynamic adaptation without requiring additional group labels. As a result, our method integrates seamlessly into standard ERM training, offering a scalable, effective solution for improving group robustness and overall model generalization.

The main \textbf{contributions} of this paper are as follows:
\begin{itemize}
\item We propose Evidential Alignment, a novel framework that integrates second-order risk minimization and evidential calibration to leverage model uncertainty for identifying and mitigating spurious correlations without group annotations.
\item We provide theoretical insights into how uncertainty quantification can capture spurious biases in latent space, offering a principled explanation for the improved robustness of our approach.
\item Extensive experiments across diverse architectures and data modalities demonstrate that Evidential Alignment significantly enhances group robustness and generalization \footnote{Our code is available at \url{https://github.com/wenqian-ye/evidential_alignment}}.
\end{itemize}

\section{Related Works}
\noindent \textbf{Spurious Correlations.}
The limitations of empirical risk minimization (ERM) in the presence of spurious correlations have been widely studied. In vision modality, ERM-trained models frequently rely on non-causal features such as background, texture, and secondary objects for classification. Similarly, in language modality, these models often exploit syntactic or statistical heuristics instead of true semantic understanding. Such reliance on spurious correlations can result in biased predictions against demographic minorities and lead to failures in high-stakes applications.

\noindent \textbf{Group Robustness Methods.} 
Existing approaches to improve group robustness against spurious correlations can be categorized based on their reliance on external supervision. Supervised methods require explicit group annotations that identify spurious correlations in the training data. If group annotations are available in the training dataset, \textit{group distributionally robust optimization} (Group DRO) \cite{sagawadistributionally} can improve group robustness by minimizing the worst-case loss across predefined groups, while other methods\cite{arjovsky2019invariant,goel2020model,zhang2022rich} learn invariant and diverse features via different techniques. Methods that only use partial group annotations include \textit{deep feature reweighting} (DFR) \cite{kirichenko2022last}, which retrains the final layer of the model on a group-balanced held-out set, and spread spurious attribute\cite{nam2022spread}, which only optimizes Group DRO with pseudo group labels. Moreover, several methods\cite{wei2023distributionally} focus on lightweight post-hoc approaches to debias the models.
Since the groups are often unknown and require expert knowledge to annotate in real-world applications, there has been great interest in methods which do not explicitly require group annotation except for model selection.
Several methods\cite{liu2021just,nam2020learning,kim2022learning} utilize auxiliary models to pseudo-label the minority group or spurious attribute. 
A recent work \cite{li2024bias} proposed an unsupervised approach that upweights misclassified samples from a bias-amplified model and selects models based on minimum class difference. While promising, this method still relies on deterministic auxiliaries for bias amplification, which may not fully capture the spurious biases. In contrast, our approach leverages uncertainty-aware information from biased models in a fully unsupervised manner. By quantifying model uncertainty and calibrating the biased model, we provide a broader perspective on identifying and mitigating spurious correlations without requiring any group annotations. 

\noindent \textbf{Uncertainty Quantification.}
Uncertainty quantification is crucial for assessing the reliability of deep learning models, particularly in safety-critical applications. Uncertainty is typically divided into aleatoric uncertainty (from inherent data noise) and epistemic uncertainty (from limited knowledge about model parameters) \cite{kendall2017uncertainties}. Methods like Bayesian Neural Networks \cite{blundell2015weight}, Monte Carlo Dropout \cite{gal2016dropout}, and Deep Ensembles \cite{lakshminarayanan2017simple} estimate uncertainty with multiple model parameters or multiple passes, which can be computationally intensive. These uncertainty quantification methods~\cite{gal2016dropout,lakshminarayanan2017simple} involve higher computational overhead and multiple model passes, thus not in our scope. Evidential deep learning \cite{sensoy2018evidential} offers an efficient yet effective alternative by modeling evidence as parameters of a Dirichlet distribution, capturing both uncertainties without the need for sampling or ensembles. While uncertainty quantification has been applied to model calibration, active learning, and out-of-distribution detection, its use in improving group robustness is less explored. Building on these insights, our method employs epistemic uncertainty to adaptively reweight samples during retraining, reducing the model's reliance on spurious attributes without requiring explicit group labels.

\begin{figure*}
    \centering
    \includegraphics[width=\linewidth]{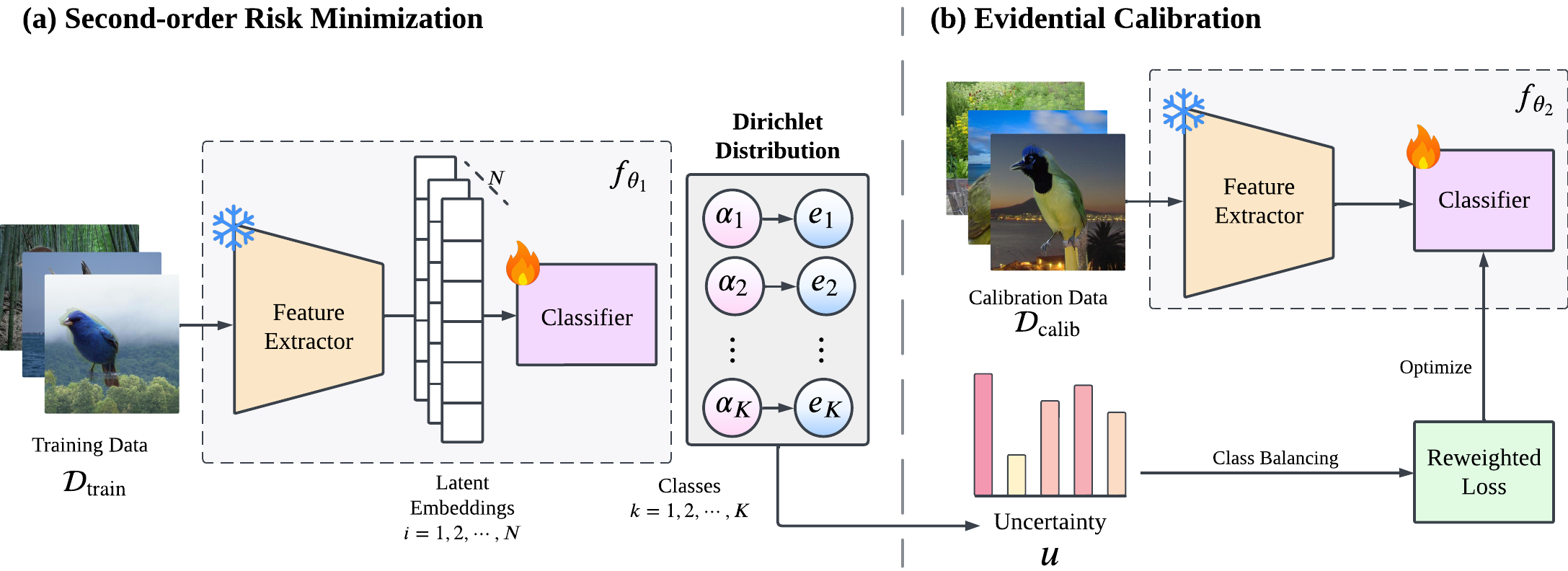}
    \caption{Overall framework of Evidential Alignment. (a) Extract the latent embeddings and train the classifier along with the Dirichlet distribution parameters using second-order risk minimization. (b) Compute the uncertainty for the calibration dataset $\mathcal{D}_{\text{calib}}$ and apply a reweighted loss with class-balanced sampling to refine the classifier.}
    \label{fig:overall_framework}
\end{figure*}
\section{Problem Formulation}
\subsection{Preliminaries}

We consider a classification problem with a training set \(\mathcal{D}_{\text{train}} = \{(x, y) | x \in \mathcal{X}, y \in \mathcal{Y}\}\), where \(x\) is a sample in the input space \(\mathcal{X}\) and \(y\) is the corresponding label in the finite label space \(\mathcal{Y}\). The training data is composed of groups \(\mathcal{D}_g^{\text{tr}}\), such that \(\mathcal{D}_{\text{train}} = \cup_{g \in \mathcal{G}} \mathcal{D}_g^{\text{tr}}\). Each group \(g := (y, a)\) is defined by a class label \(y \in \mathcal{Y}\) and a spurious attribute \(a \in \mathcal{A}\), where \(\mathcal{A}\) is the set of all spurious attributes in \(\mathcal{D}_{\text{train}}\), and \(\mathcal{G}\) is the set of all possible group labels. A group \(\mathcal{D}_g^{\text{tr}}\) consists of sample-label pairs \((x, y)\) sharing the same \(y\) and \(a\).

Our goal is to learn a model \(f_{\theta}: \mathcal{X} \rightarrow \mathcal{Y}\), parameterized by \(\theta\) from the parameter space \(\Theta\), that generalizes well across all groups on the test set \(\mathcal{D}_{\text{test}}\). The standard approach is Empirical Risk Minimization (ERM), which minimizes the expected training loss:
\begin{align}
{\mathcal{L}}_\text{ERM}(f_\theta; x,y) = \min_{\theta \in \Theta} \mathbb{E}_{(x,y) \sim \mathcal{D}_{\text{train}}} \left[ \ell(f_\theta(x), y) \right],
\end{align}
where \(\ell\) is a loss function. However, ERM-trained models often exploit spurious attributes and underperform on minority groups. To address this, we frame the group robustness problem as a min-max optimization:
\begin{align}
\min_{\theta \in \Theta} \max_{g \in \mathcal{G}} \mathbb{E}_{(x,y) \sim \mathcal{D}_g^{\text{tr}}} \left[ \ell(f_\theta(x), y) \right],
\end{align}
where \(\mathcal{D}_g^{\text{tr}}\) is the training data from group \(g\). This ensures the model performs well even on minority groups. 

\subsection{Uncertainty Decomposition}
\label{subsec:uncertainty_decomposition}
In probabilistic modeling, uncertainty is commonly divided into aleatoric (data) and epistemic (model) uncertainty. These two notions can be revealed when formulating the posterior predictive distribution of a classifier $\theta$ for a new data point $(x,y)\in \mathcal{D}$. In the following, we decompose the expression to show the foundation of our methodological design. First, we decompose uncertainty into aleatoric (data) and epistemic (model) uncertainty by marginalizing on $\theta$, i.e.,

\begin{align}
\label{eq:decompose1}
 p(y | x)=\int \underbrace{P(y | x, \theta)}_{\text {Aleatoric }} \underbrace{p(\theta | \mathcal{D})}_{\text {Epistemic }} d \theta. 
\end{align}

For a large number of real-valued parameters $\theta$ like in neural networks, this integral becomes intractable to evaluate, and thus is usually approximated using Monte Carlo samples which could incur significant computational overhead and approximation errors. To address this, a line of works \cite{malinin2018predictive,ulmer2021prior,hullermeier2021aleatoric} propose to factorize Equation \eqref{eq:decompose1} further:
\begin{align}
p(y | x)&=\iint \underbrace{P(y | \boldsymbol{\pi})}_{\text {Aleatoric }} \underbrace{p(\boldsymbol{\pi} | x, \mathcal{D})}_{\text {Distributional }} \underbrace{p(\theta | \mathcal{D})}_{\text {Epistemic }} d \boldsymbol{\pi} d \theta \\
&=\int P(y | \boldsymbol{\pi}) \underbrace{p(\boldsymbol{\pi} | x, \hat{\theta})}_{p(\theta | \mathcal{D})=\delta(\theta-\hat{\theta})} d \boldsymbol{\pi}.
\label{eq:decompose2}
\end{align}

In Equation \eqref{eq:decompose1}, we use a probabilistic distribution $\boldsymbol{\pi}$ to capture the distributional behavior of the data. Also, we can replace $p(\theta | \mathcal{D})$ with a point estimate $\hat{\theta}$ using the Dirac delta function, i.e., a single trained neural network, to get rid of the intractable integral. Although another integral remains, retrieving the uncertainty from this predictive distribution has a closed-form analytical solution for the Dirichlet distribution.

\section{Methodology}
\label{sec:method}
We propose Evidential Alignment, a two-stage approach to improve group robustness without group labels shown in Figure \ref{fig:overall_framework}. In the first stage, we extend standard ERM to a second-order risk minimization paradigm, enabling the model to estimate both predictions and their corresponding epistemic uncertainty. In the second stage, we apply an evidential calibration approach with a calibration dataset, i.e., using overconfident samples to systematically correct the model’s biases. Below, we formalize each component.

\subsection{Second-order Risk Minimization}
\label{sec:edl}
Let $(x,y) \in \mathcal{D}_{\text{train}}$ denote a training sample. Conventional first-order methods such as ERM optimize a deterministic predictive distribution $p(y | \theta,x)$ by directly estimating parameters $\theta$ (e.g., classification logits), which minimize empirical risk via objectives like negative log-likelihood. While effective for average performance, these methods conflate aleatoric (data) uncertainty and epistemic (model) uncertainty, and often exhibit brittleness under distribution shifts or spurious correlations. 

Second-order risk minimization addresses this limitation by modeling \textbf{a distribution over distributions} $p(\theta | \mathbf{\pi}(x))$, where $\mathbf{\pi}(x)$ parameterizes uncertainty about the prediction outcomes. Inspired by previous works~\cite{sensoy2018evidential,ulmer2021prior}, we extend the ERM model $f_\theta$ to output evidence values. Let the biased model parameter $\theta = (\phi_\theta,h_\theta)$ decompose into a feature extractor $\phi_\theta$ and classifier $h_\theta$.
Our goal is to learn the parameters $\theta_1 = (\phi_{\theta_1},h_{\theta_1})$ and a distribution which can jointly represent both classification objective and uncertainty quantification. 
As proved in \cite{Bergman2016SubjectiveLA},  Dirichlet distribution allows us to use the principles of evidential theory to quantify belief masses and uncertainty through a well-defined theoretical framework.
The parameters of the Dirichlet distribution $\boldsymbol{\alpha}(x) = [\alpha_1(x), \alpha_2(x), \dots, \alpha_K(x)]$ over $K$ classes are defined as:
\begin{align}
\alpha_k(x) = e_k(x) + 1, \quad \text{for } k = 1, \dots, K,
\end{align}
where $e_k \geq 0$ ($k = 1, \dots, K$) is the evidence derived for the $k$-th singleton. The non-negative evidence \(e_k(x)\) for each class \(k \in \{1,\dots,K\}\) is derived from the model outputs from classifier $h_{\theta_1}$.  The Dirichlet distribution over the class probabilities, that is, $\mathbf{p}(x) = [p_1(x), p_2(x), \dots, p_K(x)]$,
can be expressed as:
\begin{align}
\text{Dir}(\mathbf{p}(x) \, | \, \boldsymbol{\alpha}(x)) = \frac{1}{B(\boldsymbol{\alpha}(x))} \prod_{k=1}^K p_k(x)^{\alpha_k(x) - 1},
\end{align}
where \(B(\boldsymbol{\alpha}(x))\) is the multivariate Beta function. The expected class probabilities are:
\begin{align}
\mathbb{E}[p_k(x)] = \frac{\alpha_k(x)}{S(x)}, \quad \text{where } S(x) = \sum_{k=1}^K \alpha_k(x).
\end{align}

By analyzing both the prediction results and the corresponding uncertainty estimates, we can gain valuable insights on how the model is influenced by the training data distribution. For a classification task with \(K\) classes, the model outputs non-negative evidence values \(e_k(x) \geq 0\) for each class \(k\). 

To guide the learning of both prediction and the parametrization of Dirichlet distribution, we formulate the loss function for \textbf{second-order risk minimization} as:
\begin{align}
\mathcal{L}_1(x, y | \theta_1) = \underbrace{-\log ( \mathbb{E}[p_{y}(x)] )}_{\text{Classification Objective}} + \lambda \cdot \underbrace{\text{KL}( \text{Dir}( \boldsymbol{\alpha}(x)) \, \| \, \text{Dir}( \mathbf{1}))}_{\text{Evidence Regularization}},
\end{align}
where \(\lambda\) is an evidence regularization coefficient and \(\text{Dir}( \mathbf{1} )\) represents a non-informative Dirichlet prior with all concentration parameters equal to 1. The Evidence Regularization term penalizes overconfidence and promotes uncertainty where appropriate. We first define the classification objective as follows.

\begin{align}
\mathcal{L}_{\text{cls}}(x, y | \theta_1) = -\log ( \frac{\alpha_{y}(x)}{S(x)} ) = -\log ( \frac{e_{y}(x) + 1}{\sum_{k=1}^K ( e_k(x) + 1 )} ).
\end{align}

For the Evidence Regularization term, it controls overconfidence by penalizing large deviations from the uniform prior. It is defined as the Kullback-Leibler (KL) divergence between the predicted Dirichlet distribution and a non-informative prior (a uniform Dirichlet distribution).

\begin{align}
\text{KL}( \text{Dir}(\boldsymbol{\alpha}) \, \| \, \text{Dir}(\mathbf{1}) ) = \log ( \frac{B(\mathbf{1})}{B(\boldsymbol{\alpha})} ) + \sum_{k=1}^K (\alpha_k - 1) ( \psi(\alpha_k) - \psi(S) ),
\end{align}
where \(\psi(\cdot) = \frac{d}{dz} \ln \Gamma(\cdot)\) is the digamma function. The loss for a sample $(x,y) \in \mathcal{D}_{\text{train}}$ becomes:

\begin{align}
 \mathcal{L}_1(x, y | \theta_1) = -\log ( \frac{e_{y}(x) + 1}{\sum_{k=1}^K ( e_k(x) + 1)}) + \lambda_t \cdot \text{KL}( \text{Dir}( \boldsymbol{\alpha}(x) ) \, \| \, \text{Dir}( \mathbf{1} )).
\label{eq:total_loss}   
\end{align}

In practice, we adopt a dynamic evidence regularization coefficient $\lambda_t = \min (\frac{t}{\eta}, 1)$ to stabilize the training process, where $t$ is the current epoch number and $\eta$ is a hyperparameter of the annealing step.

\subsection{Evidential Calibration}

After training the model with the second-order risk, we use the estimated epistemic uncertainty to compute a weight for each sample \((x,y)\) in the calibration set \(\mathcal{D}_{\text{calib}}\), then calibrate the biased model. As shown in Equation~\eqref{eq:decompose2}, the predictive distribution can be expressed via \(p(y | \boldsymbol{\pi})\) and \(p(\boldsymbol{\pi} | x,\hat{\theta})\). In this stage, we aim to learn unbiased representation based on the 
estimated epistemic uncertainty. 

Firstly, we can compute the uncertainty estimate $u(x)$ from the learned Dirichlet distribution.
\begin{align}
u(x) 
= 
\frac{K}{S(x)} 
= 
\frac{K}{\sum_{k=1}^K \alpha_k(x)},
\end{align}
where \(S(x)\) is the sum of the Dirichlet parameters from the learned evidential model. Higher values of \(u(x)\) suggest larger epistemic uncertainty in the prediction for sample \(x\). To perform Bayesian model averaging (BMA) during retraining, we integrate the estimated posterior probabilities by reweighting the loss contributions with class-balanced samples. 

We express the reweight function \(w(x,y)\) in the calibration dataset \(\mathcal{D}_{\mathrm{calib}}\):
\begin{align}
w(x,y)
=
\mathbbm{1}\bigl[f_{\theta_1}(x) = y\bigr]
+
u(x) \cdot \mathbbm{1}\bigl[f_{\theta_1}(x) \neq y\bigr],
\end{align}
where $\mathbbm{1}[\cdot]$ is the indicator function. This upweights misclassified samples proportionally to their epistemic uncertainty. The objective function for retraining the model parameters \(\theta_2\) over the calibration dataset $\mathcal{D}_{\text{calib}}$ becomes:
\begin{align}
\label{eq:evidence-calibration}
\mathcal{L}_2(\mathcal{D}_{\text{calib}} | u,\theta_2)
= 
\mathop{\mathbb{E}}\limits_{(x,y)\sim\mathcal{D}_{\text{calib}}}
\bigl[w(x,y) 
\, \ell_{\text{ce}}\bigl(f_{\theta_2}(x), y\bigr) \bigr]
\;+\; 
\beta \, \bigl\lVert \theta_2 - \theta_1 \bigr\rVert_{2}^2,
\end{align}
where \(\ell_{\text{ce}}(f_\theta(x), y)\) is the cross-entropy loss function for sample \((x,y)\). Here, \(\theta_1\) denotes the parameters learned at the end of second-order risk minimization, and \(\beta\) controls how strongly we regularize the retraining objective. During training, we draw samples from \(\mathcal{D}_{\text{calib}}\) in a class-balanced manner.

In practice, we only retrain the last layer of the model to learn the invariant features like previous works~\cite{kirichenko2022last,labonte2024towards}. This approach enhances the model's robustness to spurious correlations and improves generalization across all groups without requiring explicit group labels.

\begin{algorithm}[ht]
\caption{\textit{Evidential Alignment}}
\label{algo:evidential}
\begin{algorithmic}[1]
    \State \textbf{Input:} Training dataset $\mathcal{D}_\text{train}$, calibration dataset $\mathcal{D}_\text{calib}$, biased model $f_{\theta}$, number of epochs $T_1$, $T_2$, KL regularization term $\lambda$
    \State \textbf{Output:} Updated model parameters $\theta^*$
    \State \textit{\textbf{Stage 1: Second-order Risk Minimization}}
    \For{epoch $t = 1$ to $T_1$}
        \For{each sample $(x, y) \in \mathcal{D}_\text{train}$}
            \State Estimate evidence $e_k(x)$ for each class $k$
            \State Calculate Dirichlet parameters: $\alpha_k(x) = e_k(x) + 1$
            \State Calculate expected class probabilities: $\mathbb{E}[p_k(x)]$
            \State Calculate classification loss: $\mathcal{L}_{\text{cls}}(x, y | \theta_1)$
            \State Calculate KL Divergence: $\text{KL}(\text{Dir}(\boldsymbol{\alpha}(x)) \,\|\, \text{Dir}(\mathbf{1}))$
            \State Optimize last-layer parameters $\theta_1$ with $\mathcal{L}_1(x, y | \theta_1)$
        \EndFor
    \EndFor
    \State \textit{\textbf{Stage 2: Evidential Calibration}}
    \For{epoch $t = 1$ to $T_2$}
        \For{each sample $(x, y) \in \mathcal{D}_\text{calib}$}
            \State Calculate uncertainty: $u(x) = \frac{K}{S(x)}$
            \State Calculate weights: $w(x,y)$
            \State Calculate reweight loss: $w(x,y) \cdot \ell(f_\theta(x), y)$
            \State Optimize last-layer parameters $\theta_2$ with 
        \(
        \mathcal{L}_2(\mathcal{D}_{\text{calib}} | u,\theta_2)
        \)
        \EndFor
    \EndFor
    \State \textbf{Return:} Updated model parameters $\theta^*$
\end{algorithmic}
\end{algorithm}

\subsection{Overview of Evidential Alignment}
We summarize Evidential Alignment in Algorithm~\ref{algo:evidential}. It consists of two stages. The first stage performs second-order risk minimization, where the model estimates uncertainty using a Dirichlet distribution and incorporates a KL regularization. The second stage uses the estimated uncertainty to compute sample weights on a calibration set and retrains the last layer with reweighted loss.

\subsection{Theoretical Analysis}
In this section, we provide theoretical guarantees for our method, demonstrating that (1) second-order risk minimization converges to a Dirichlet distribution of the ERM model, and (2) the PAC-Bayes generalization bound for the reweighted empirical risk holds. 

\begin{theorem}[Evidence Lower Bound for Second-Order Risk]
\label{thm:elbo}
    Let \( p(\boldsymbol{\pi} | y) \) be the true posterior distribution over class probabilities given label \( y \), and let \( p(\boldsymbol{\pi} | x, \theta) \) be the variational evidential distribution approximating it, where \( \theta \) represents the model parameters. Then, training with the second-order risk minimization optimizes the evidence lower bound (ELBO), which maximizes the log marginal likelihood of observed labels and learns evidence on the predictions. Formally, the loss function:
    \begin{equation}
        \mathcal{L}_{\mathrm{ELBO}} = \mathbb{E}_{p(\boldsymbol{\pi} | x, \theta)}[-\log P(y | \boldsymbol{\pi})] + \operatorname{KL}[p(\boldsymbol{\pi} | x, \theta) \,\|\, p(\boldsymbol{\pi})]
    \end{equation}
    is an upper bound on the negative log-marginal likelihood \( -\log p(y) \).
\end{theorem}
This result informs the core principle of second-order risk minimization: by modeling a Dirichlet distribution for each sample, we impose a Bayesian perspective that naturally captures model confidence and uncertainty.

\begin{theorem}[PAC-Bayes Bound for Reweighted Empirical Risk]
\label{thm:pacbayes-worstgroup}
Let $\mathcal{H}$ be a hypothesis space with $[0,1]$-bounded loss, and let $\mathcal{G}=\{1,\dots,|\mathcal{G}|\}$ be a finite set of latent groups. Suppose we have $n_g$ i.i.d.\ training samples from group $g \in \mathcal{G}$, and define $n_{\min}=\min_{g} n_g$. Let $R_g(f)$ be the true risk of hypothesis $f \in \mathcal{H}$ on group $g$, and let $\hat{R}_g(f)$ be the empirical risk of $f$ on the $n_g$ training samples in group $g$. Suppose a (potentially data-dependent) weighting scheme $\{w_g\}_{g\in\mathcal{G}}$ is chosen such that $w_g \ge 0$ and $\sum_{g=1}^{|\mathcal{G}|} w_g = 1$. Define the reweighted empirical risk
\begin{align}
\hat{R}_w(f)
\;=\;
\sum_{g=1}^{|\mathcal{G}|} 
w_g
\;\hat{R}_g(f),
\end{align}
and let $\alpha = \min_{g}\,w_g$. Let $P$ be a prior distribution over $\mathcal{H}$, and let $Q$ be a (posterior) distribution over $\mathcal{H}$. Then, for any $0 < \delta < 1$, with probability at least $1-\delta$ over the random draw of all training samples, the worst-group risk of $Q$ satisfies:
\begin{align}
\max_{1 \le g \le |\mathcal{G}|}\,
\mathbb{E}_{f \sim Q}\bigl[R_g(f)\bigr]
\le
\frac{1}{\alpha}
\mathbb{E}_{f \sim Q}\bigl[\hat{R}_w(f)\bigr]
+
\sqrt{
  \frac{
    \mathrm{KL}\bigl(Q\,\|\,P\bigr)
    + 
    \ln \bigl(\tfrac{|\mathcal{G}|}{\delta}\bigr)
  }
  {2\,n_{\min}}
}.
\end{align}
\end{theorem}

This theorem shows that if a reweighting scheme keeps the reweighted empirical risk low, the worst-group risk remains bounded with high probability. Consequently, optimizing the reweighted empirical risk with a suitably chosen weighting scheme that accounts for the worst-case group leads to better group robustness. 

We defer the detailed proofs of both Theorem \ref{thm:elbo} and Theorem \ref{thm:pacbayes-worstgroup} to the Appendix.

\section{Experiments}
We evaluate Evidential Alignment on a range of group robustness benchmarks and provide ablation studies and additional results to further support our motivation and methodology designs.

\subsection{Datasets}
We evaluate six benchmark datasets for group robustness in vision and language tasks. We also move beyond simple binary classification tasks that have been the focus of most recent works for studying spurious correlations. Below, we show the dataset details. 

\noindent \textbf{Colored MNIST}~\cite{arjovsky2019invariant} is a variant of the MNIST dataset designed for image classification, where digit colors introduce spurious correlations with class labels. In our setup, we choose two classes: digit 1 as class 0 and digit 8 as class 1 with two colors green and red as spurious attributes, following ~\citet{arjovsky2019invariant}. 

\noindent \textbf{Waterbirds}~\cite{sagawadistributionally} is an image classification dataset with waterbird and landbird classes selected from the CUB dataset~\cite{WelinderEtal2010}. The spurious attribute is the water or land backgrounds from the Places dataset \cite{zhou2017places}, where more landbirds are present on land backgrounds than waterbirds, and vice versa.

\noindent \textbf{CelebA}~\cite{liu2015deep} is a large scale image dataset with celebrity faces. The task is to identify whether the celebrity's hair color (blonde or non-blonde). The spurious attribute is gender (male and feamale) where the majority group is blonde women in the training set.

\noindent \textbf{CheXpert}~\cite{irvin2019chexpert} is a large scale clinical dataset of chest radiograph. The dataset includes six spurious attributes, defined by the combination of race (White, Black, Other) and gender (Male, Female). The classification task involves predicting two diagnostic outcomes: “No Finding” (positive) or “Finding” (negative).

\noindent \textbf{MultiNLI}~\cite{williams2017broad} is a large-scale natural language inference dataset that contains sentence pairs across multiple genres, annotated with entailment classes (entailment, contradiction, and neutral). The spurious attribute involves syntactic or lexical patterns (e.g., negation words) that correlate with specific labels.

\noindent \textbf{CivilComments}~\cite{borkan2019nuanced} is a binary text classification dataset sourced from internet comments. The task is to predict whether a comment contains toxic language. The spurious attributes include eight demographic identities (e.g., gender, race, etc) that correlate with the label. The dataset uses standard splits from the WILDS benchmark~\cite{koh2021wilds}.

\begin{table}[ht]
    \caption{Comparison of Test Accuracy on CMNIST between ERM and Ours (\%). (Class 1, Color 0) is the minority group.}
    \label{tab:second_table}
    \centering
    \begin{tabular}{lcc}
        \toprule
        \textbf{Accuracy} & \textbf{ERM} & \textbf{Ours} \\ \midrule
        Average            & 76.34       & 96.21        \\ \midrule
        Class 0, Color 0   & 100.00      & 100.00        \\
        Class 0, Color 1   & 82.81       & 97.83        \\
        Class 1, Color 0   & 3.74        & 84.58        \\
        Class 1, Color 1   & 100.00      & 99.45        \\ \midrule
        WGA                & 3.74        & 84.58        \\ \bottomrule
    \end{tabular}
\end{table}



\subsection{Experimental Setups}
\noindent \textbf{Training Details.}
We first train the ERM base models with pretrained weights. For the image datasets, we use a ResNet-50~\cite{he2016deep} model pretrained on ImageNet as the backbone. For the text datasets, we use the BERT-base-uncased model~\cite{kenton2019bert} pretrained on the BookCorpus and English Wikipedia. Consistent with the approach in \cite{yang2023change}, we don't apply any data augmentations during ERM training. For our method, we constructed the calibration dataset $\mathcal{D}_{\text{calib}}$ using the same half of the validation set and used it for both second-order minimization and evidential calibration. The remaining half of the validation set is only used for model selection and hyperparameter tuning. During both second-order risk minimization and evidential calibration, we only train the last-layers of the models. We ran the training under three different random seeds and reported average accuracies along with standard deviations. All experiments were conducted on NVIDIA A6000 GPUs. We provide full experimental details in the Appendix.

\begin{table*}[ht]
\caption{Comparison of worst-group accuracy (WGA), average accuracy (Acc.), and accuracy gap across image datasets. Best worst-group accuracies are highlighted in \textbf{boldface}. $^\dagger$ denotes using a fraction of validation data for model fine-tuning.  }
\label{tab:image}
\centering
\begin{tabular}{lccccccccc}
\toprule
\multirow{2}{*}{\textbf{Method}} & \multirow{2}{*}{\textbf{Backbone}} & \multicolumn{2}{c}{\textbf{Group Annotations}} & \multicolumn{3}{c}{\textbf{Waterbirds}} & \multicolumn{3}{c}{\textbf{CelebA}} \\
\cmidrule(lr){3-4} \cmidrule(lr){5-7} \cmidrule(lr){8-10}
& & Train & Val & WGA(\%) \(\uparrow\) & Acc.(\%) \(\uparrow\) & Gap (\%) \(\downarrow\) & WGA(\%) \(\uparrow\) & Acc.(\%) \(\uparrow\) & Gap(\%) \(\downarrow\) \\
\midrule
ERM~\cite{vapnik1999overview}         &  ResNet50  & - & -  & 72.6 & 97.3 & 24.7 & 47.2 & 95.6 & 48.4 \\ \midrule
JTT~\cite{liu2021just}         &  ResNet50  & No & Yes  & 86.7 & 93.3 & 6.6 & 81.1 & 88.0 & 6.9 \\ 
CnC~\cite{zhang_correct-n-contrast_2022}         &  ResNet50  & No & Yes  & 88.5$_{\pm 0.3}$ & 90.9$_{\pm 0.1}$ & 2.4 & 88.8$_{\pm 0.9}$ & 89.9$_{\pm 0.5}$ & 1.1 \\ 
AFR~\cite{qiu2023simple} & ResNet50 & No & Yes & 90.4$_{\pm 1.1}$ & 94.2$_{\pm 1.2}$ & 3.8 & 82.0$_{\pm 0.5}$ & 91.3$_{\pm 0.3}$ & 9.3 \\ 
DFR$^\dagger$~\cite{kirichenko2022last} & ResNet50 & No & Yes & 92.9$_{\pm 0.9}$ & 94.2$_{\pm 0.3}$ & 2.5 & 88.3$_{\pm 1.1}$ & 91.3$_{\pm 0.5}$ & 3.0 \\ 
SELF$^\dagger$~\cite{labonte2024towards} & ResNet50  & No & Yes  & 93.0$_{\pm 0.3}$ & 94.0$_{\pm 1.7}$ & 1.0 & 83.9$_{\pm 0.9}$ & 91.7$_{\pm 0.4}$ & 7.8 \\ 
\midrule
LfF~\cite{nam2020learning}         &  ResNet50  & No & No  & 78.0 & 91.2 & 13.2 & 77.2 & 85.1 & 7.9 \\ 
BPA     \cite{seo2022unsupervised}      & ResNet50                                 & No                 & No               & 71.4              & -                 & -                       & 82.5                                & -                 & -                       \\
GEORGE \cite{sohoni2020no}   & ResNet50                                     & No                 & No               & 76.2              & 95.7              &    19.5                     & 52.4                                & 94.8              &                  42.4       \\
BAM    \cite{li2024bias}              & ResNet50             & No                 & No               & 89.1$_{\pm 0.2}$ & 91.4$_{\pm 0.3}$ &      2.3                   & 80.1$_{\pm 3.3}$ & 88.4$_{\pm 2.3}$ &             8.3            \\ 
\grayrow \textbf{Ours$^\dagger$} & ResNet50 & No & No & \textbf{92.2}$_{\pm 0.7}$ & 95.1$_{\pm 0.5}$ & 2.9 & \textbf{84.4}$_{\pm 0.8}$ & 92.3$_{\pm 0.6}$ & 7.9 \\ 
\bottomrule
\end{tabular}%
\end{table*}

\noindent \textbf{Model Selection Strategy.}
We tested three model selection strategies in this work. Using the second half of the validation set, we compute (1) standard average accuracy, which represents the overall classification accuracy across all validation samples, (2) worst-class accuracy, which measures the accuracy on the least accurate class to assess class-wise performance disparities, and (3) worst-group accuracy (WGA), which evaluates the model’s performance on the most challenging group, providing insight into its robustness against spurious correlations. In practice, we use worst-class accuracy to make fair comparisons where there are no group labels available.

\noindent \textbf{Evaluation Metric.}
We use the worst-group accuracy (WGA) as the \textbf{main metric} for evaluating group robustness, shown in Equation~\eqref{eq:wga}. WGA computes the accuracy of test samples that contain the spurious attribute associated with the other class during training. It measures the model’s accuracy on the group with the lowest performance. 
\begin{align}
\label{eq:wga}
\operatorname{WGA}(f_\theta; x, y) := \min_{g \in \mathcal{G}} \mathbb{E}_{(x,y) \sim \mathcal{D}_g} [ \mathbbm{1}[f_\theta(x) = y]],
\end{align}
where \(\mathbbm{1}[f_\theta(x) \neq y]\) is the 0-1 loss. In addition, we also measure the \textit{accuracy gap} between standard average accuracy and worst-group accuracy as a measure of the robustness of the models. A high worst-group accuracy and a low accuracy gap indicate that the classifier is robust to spurious correlations and can fairly predict samples from different groups.

\subsection{Synthetic Experiment}

To demonstrate the concept of Evidential Alignment, we first conducted a simple synthetic experiment with the Colored MNIST dataset~\cite{arjovsky2019invariant}. In this experiment, we manually introduced spurious correlations between the color and the class label, with 90\% of the training samples (\(p_{\text{corr}} = 0.9\)) exhibiting this color-class association, while only 10\% of the test samples (\(p_{\text{corr}} = 0.1\)) retained the same correlation. For Class 0, there are 6,398 red instances and 344 green instances. For Class 1, there are 325 red instances and 5,526 green instances. Notably, the (0, green) and (1, red) represent the minority groups in this synthetic dataset. 

We used Evidential Alignment to learn group robustness models. First, we trained a simple CNN model, LeNet-5~\cite{lecun1998gradient} using ERM with second-order risk minimization for 10 epochs. After obtaining the uncertainty estimates from the ERM model, we calculated the weights and optimize the evidential calibration objective to mitigate the impact of the spurious correlations.

\begin{figure}[h]
    \centering
    \includegraphics[width=\linewidth]{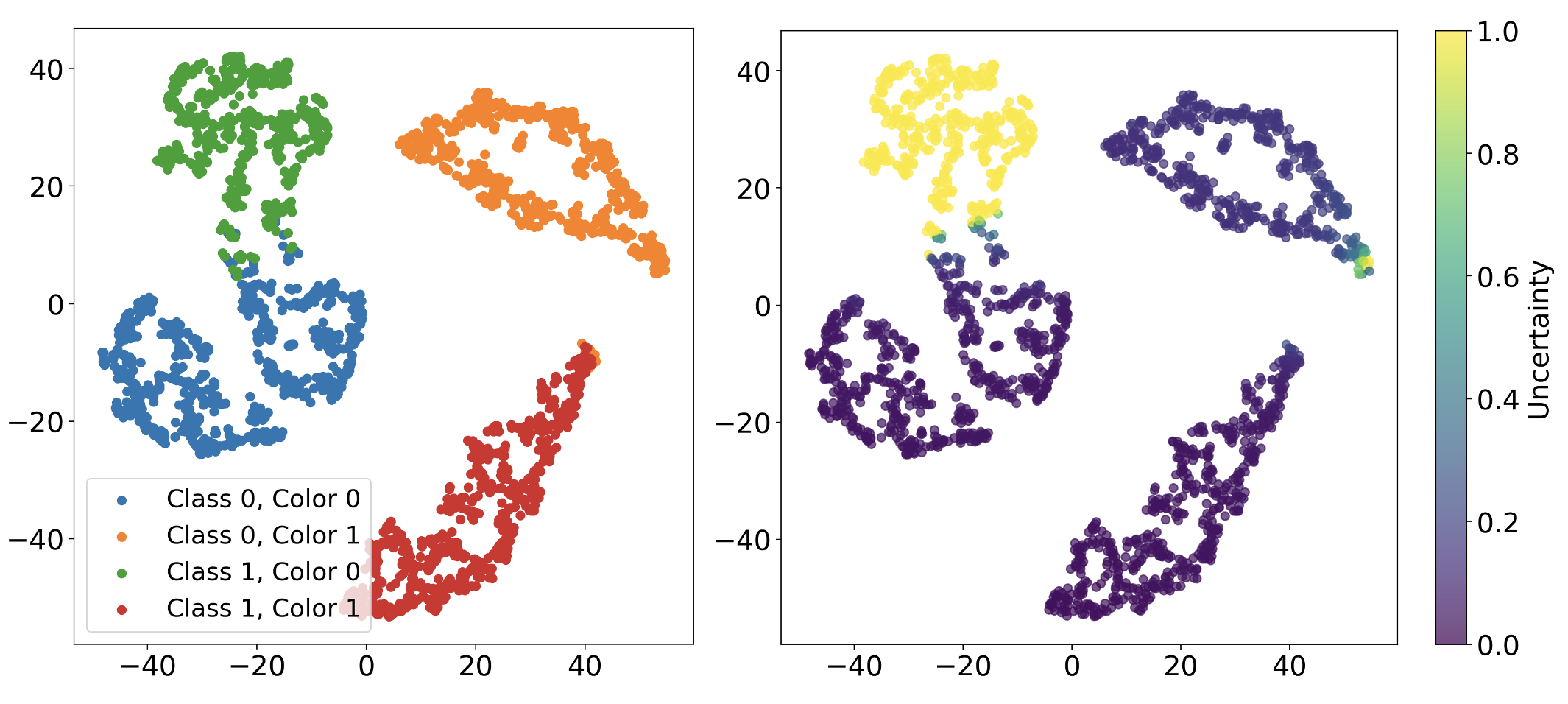}
    \caption{t-SNE visualization of the embeddings on the test set with ERM model. Left: test data samples in four groups. Right: quantified uncertainty for each sample (Yellow denotes high uncertainty; Dark purple denotes low uncertainty). }
    \label{fig:TSNE}
\end{figure}

We first show that second-order risk minimization can give a good estimate of challenging samples (in minority groups) in Figure~\ref{fig:TSNE} with t-SNE visualization~\cite{van2008visualizing}. As shown in Table~\ref{tab:second_table}, this approach significantly improves the performance for the minority group (Class 1, Color 0), with accuracy increasing from 3.74\% to 84.58\%. The performance of the majority groups remains unaffected by the retraining process. Here, the purpose of selecting digits 1 and 8 in this experiment was only to provide a clear contrast between the two classes. Other digit pairs, such as 0 and 7, would yield similar results.

\begin{table*}[ht]
\caption{Comparison of worst-group accuracy (WGA), average accuracy (Acc.), and accuracy gap across text datasets. Best worst-group accuracies are highlighted in \textbf{boldface}. $^\dagger$ denotes using a fraction of validation data for model fine-tuning. }
\label{tab:text}
\centering
\begin{tabular}{lccccccccc}
\toprule
\multirow{2}{*}{\textbf{Method}} & \multirow{2}{*}{\textbf{Backbone}} & \multicolumn{2}{c}{\textbf{Group Annotations}} & \multicolumn{3}{c}{\textbf{MultiNLI}} & \multicolumn{3}{c}{\textbf{CivilComments}} \\
\cmidrule(lr){3-4} \cmidrule(lr){5-7} \cmidrule(lr){8-10}
& & Train & Val & WGA(\%) \(\uparrow\) & Acc.(\%) \(\uparrow\) & Gap(\%) \(\downarrow\) & WGA(\%) \(\uparrow\) & Acc.(\%) \(\uparrow\) & Gap(\%) \(\downarrow\) \\
\midrule
ERM~\cite{vapnik1999overview}         &  BERT  & - & -  & 67.9 & 82.4 & 14.5 & 57.4 & 92.6 & 35.2 \\ \midrule
JTT~\cite{liu2021just}         &  BERT  & No & Yes  & 72.6 & 78.6 & 6.0 & 69.3 & 91.1 & 21.8 \\ 
CNC~\cite{ zhang_correct-n-contrast_2022} & BERT & No & Yes & - & - & - & 68.9$_{\pm 2.1}$ & 81.7$_{\pm 0.5}$ & 12.8 \\ 
BAM~\cite{li2024bias} & BERT & No & Yes & 71.2$_{\pm 1.6}$ & 79.6$_{\pm 1.1}$ & 8.4 & 79.3$_{\pm 2.7}$ & 88.3$_{\pm 0.8}$ & 9.0 \\ 
AFR~\cite{qiu2023simple} & BERT & No & Yes & 73.4$_{\pm 0.6}$ & 81.4$_{\pm 0.2}$ & 8.0 & 68.7$_{\pm 0.6}$ & 89.8$_{\pm 0.6}$ & 21.1 \\ 
DFR$^\dagger$~\cite{kirichenko2022last} & BERT & No & Yes & 70.8$_{\pm 0.8}$ & 81.7$_{\pm 0.2}$ & 10.9 & 81.8$_{\pm 1.6}$ & 87.5$_{\pm 0.2}$ & 5.7 \\ 
SELF$^\dagger$~\cite{labonte2024towards} &  BERT  & No & Yes & 70.7$_{\pm 2.5}$ & 81.2$_{\pm 0.7}$ & 10.5 & 79.1$_{\pm 2.1}$ & 87.7$_{\pm 0.6}$ & 8.6 \\ 
\midrule
LfF~\cite{nam2020learning}         &  BERT  & No & No  & 70.2 & 80.8 & 10.6 & 58.8 & 92.5 & 33.7 \\ 
BAM~\cite{li2024bias} & BERT & No & No & 70.8$_{\pm 1.5}$ & 80.3$_{\pm 1.0}$ & 9.5 & 79.3$_{\pm 2.7}$ & 88.3$_{\pm 0.8}$ & 9.0 \\ 
\grayrow \textbf{Ours$^\dagger$} & BERT & No & No & \textbf{74.5}$_{\pm 1.2}$ & 80.6$_{\pm 0.8}$ & 7.3 & \textbf{80.2}$_{\pm 0.9}$ & 87.1$_{\pm 0.4}$ & 6.9 \\ 
\bottomrule
\end{tabular}%

\end{table*}

\begin{table}[t]
\caption{Worst-group (WGA) and average accuracy (Acc.) on the CheXpert dataset. All group robustness methods use the same half of the validation set. Best worst-group accuracies and average accuracies are highlighted in \textbf{boldface}.}
\label{tab:chexpert}
\centering
\begin{tabular}{cccc}
\toprule
\textbf{Method} & \textbf{Backbone} & WGA(\%) $\uparrow$ & Acc.(\%) $\uparrow$  \\ 
\midrule
ERM \cite{vapnik1999overview} & ResNet50      & 22.0$_{\pm1.6}$  & 90.8$_{\pm0.1}$   \\
\midrule
JTT \cite{liu2021just}   & ResNet50          & 60.4$_{\pm4.9}$  & 75.2$_{\pm0.8}$   \\
DFR \cite{kirichenko2022last}   & ResNet50   & 72.7$_{\pm1.5}$  & 78.7$_{\pm0.4}$   \\
AFR \cite{qiu2023simple}      & ResNet50     & 72.4$_{\pm2.0}$  & 76.8$_{\pm1.1}$   \\
\grayrow
\textbf{Ours} & ResNet50 & \textbf{73.6}$_{\pm1.1}$ & \textbf{79.9}$_{\pm0.9}$  \\
\bottomrule
\end{tabular}
\end{table}

\subsection{Main Results}

In this section, we compare our method, Evidential Alignment, with state-of-the-art baseline methods for improving group robustness, and in particular, with those not requiring group annotations in both training and validation sets. We tested our method on real-world datasets, covering both image and text modalities. We are especially interested in methods that can achieve a high worst-group accuracy (WGA) as well as a low accuracy gap, as such methods demonstrate high group robustness and do not exhibit strong prediction biases towards certain data groups.

As shown in the bottom parts of Tables~\ref{tab:image} and~\ref{tab:text}, which include methods that do not use group annotations, our method achieves the best WGA, demonstrating its superior capability in improving a model's group robustness without explicit supervision with group labels. For methods that require group labels for model selection, such as JTT~\cite{liu2021just} and AFR~\cite{qiu2023simple}, our method consistently outperforms them in terms of WGA and accuracy gap. We note that CnC~\cite{ zhang_correct-n-contrast_2022} achieves the best WGA on CelebA; however, unlike our method, it fails to achieve superior performance on the other datasets. Similar to our method, SELF~\cite{labonte2024towards} uses a half of the validation data for training without group labels. Our method outperforms SELF on three out four datasets. DFR~\cite{kirichenko2022last}, which uses group labels in a half of validation data for training, is expected to achieve high WGAs on these datasets. We show that, even without group labels, our method is competitive with DFR and outperforms it on the MultiNLI dataset. In Table~\ref{tab:chexpert}, we used the CheXpert dataset, which contains chest radiograph confounded with gender and race, as an example to further demonstrate the effectiveness of our method on critical domains. Our method achieves both the best WGA and average accuracy on this dataset, exhibiting strong robustness against spurious attributes. 

Notably, our method introduces minimal computational overhead compared to existing approaches such as JTT~\cite{liu2021just} and CnC~\cite{ zhang_correct-n-contrast_2022}. Unlike our method, JTT and CnC require retraining an additional model from scratch. In contrast, our method, which consists of a standard ERM training with evidence regularization and an evidential calibration, only manipulates the last-layer of the model.  Thus, our approach is efficient in terms of both computation and memory.

\begin{figure*}[h]
    \centering
    \includegraphics[width=0.8\linewidth]{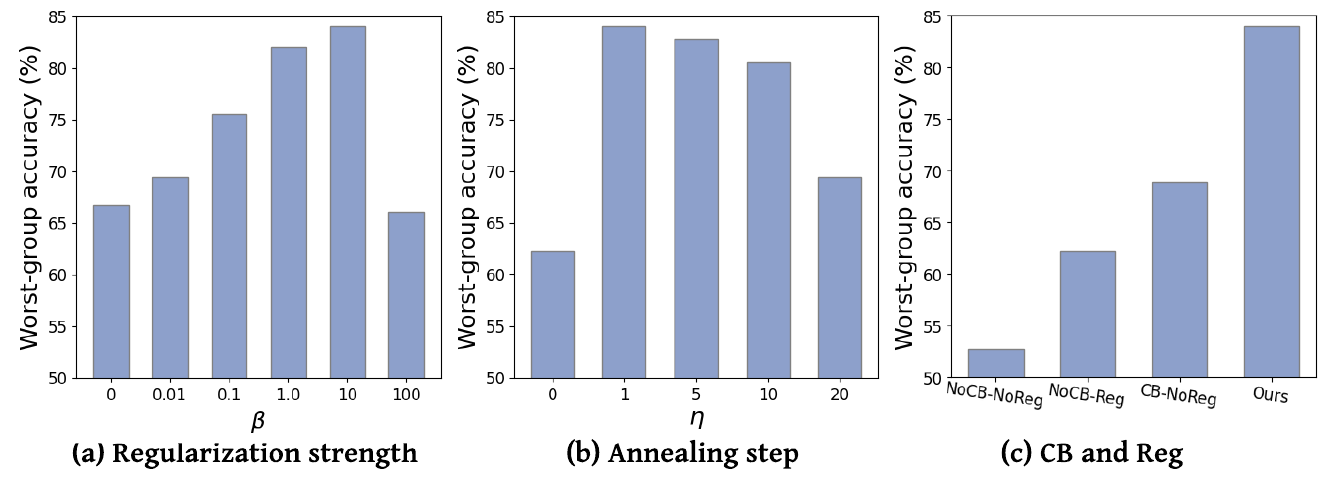}
    \caption{Ablation study on how (a) regularization strength $\beta$, (b) annealing step $\eta$, and (c) class balancing (CB) and regularization (Reg) affect our method's worst-group accuracy on the CelebA dataset.}
    \label{fig:ablation}
\end{figure*}
\subsection{Ablation Studies}
In Figure~\ref{fig:ablation}, we present the effect of the annealing step $\eta$ (from Equation \eqref{eq:total_loss}) and the regularization strength $\beta$ (from Equation \eqref{eq:evidence-calibration}), as well as how class balancing and our proposed KL (from Equation \eqref{eq:total_loss}) and weight (from Equation \eqref{eq:evidence-calibration}) regularization contribute to group robustness. 

Specifically, Figure~\ref{fig:ablation}(a) highlights the effectiveness of our proposed weight regularization when $\beta>0$. Compared with no weight regularization when $\beta=0$, our method benefits with a moderate $\beta$, e.g., $\beta=10$ on the CelebA dataset. Figure~\ref{fig:ablation}(b) demonstrates the importance of KL regularization as well as how it is scheduled in the second-order risk minimization. We denote no such regularization with $\eta=0$. When $\eta=20$, which equals the total number of training epochs, the regularization strength is progressively increased until the end of training. When  $\eta<20$, the regularization strength peaks at the $\eta$'th epoch and remains unchanged for the rest of training. In general, adding the KL regularization benefits finding good uncertainty estimations and thereby improves a model's group robustness. Selecting an appropriate $\eta$ is dataset-dependent, and in practice, a small $\eta$ is favored for datasets with severe group imbalances, such as CelebA. Figure~\ref{fig:ablation}(c) analyzes how class balancing (CB) and our proposed regularizations (Reg) benefit from each other to achieve the optimal group robustness. Compared with NoCB-NoReg, which is the vanilla evidential calibration algorithm, adopting CB (denoted as CB-NoReg) improves group robustness to some extend. Our method, which incorporates both CB and Reg, provides better uncertainty estimations than NoCB-Reg and CB-NoReg and significantly boosts models' group robustness by achieving the highest worst-group accuracy.

\begin{table}[ht]
\caption{Comparison of other reweighting-based methods on computational complexity. 
}
\label{tab:complexity}
\centering
\begin{tabular}{lc}
\toprule
\textbf{Method} &  \textbf{Time Complexity} \\
\midrule
JTT  &  \(O(N\,T\,\xi_{\text{full}})\) \\
DFR  & \(O(N\,T_2\,\xi_{\text{last}})\) \\
AFR  & \(O(N\,T_2\,\xi_{\text{last}} + N\,\xi_{fw})\) \\
\grayrow
\textbf{Ours} & \(O(N\,T_1\,\xi_{\text{last}} + N\,T_2\,\xi_{\text{last}})\) \\
\bottomrule
\end{tabular}
\end{table}

\subsection{Complexity Analysis}
\label{sec:complexity}
We analyze the computational complexity of our method, Evidential Alignment, compared with other reweighting-based approaches such as JTT, DFR, and AFR (see Table~\ref{tab:complexity}). Let \(N\) denote the number of training samples. \(T\) is the number of epochs for full model training (used by JTT). \(T_1\) is the number of epochs for the second-order risk minimization stage in Evidential Alignment. \(T_2\) is the number of epochs for retraining only the last layer (used by DFR, AFR, and Evidential Alignment). We use \(\xi_{\text{full}}\) for the cost per sample per epoch in full model training and \(\xi_{\text{last}}\) for last-layer training, where \(\xi_{\text{full}} \gg \xi_{\text{last}}\). For methods that train the full model over \(T\) epochs (i.e., JTT), the complexity is \(O(N\,T\,\xi_{\text{full}})\). This induces the highest cost. In contrast, DFR retrains only the last layer for \(T_2\) epochs, yielding a complexity of \(O(N\,T_2\,\xi_{\text{last}})\). Note that DFR requires group annotations. AFR extends DFR by precomputing sample losses. Let \(\xi_{fw}\) be the cost for a single feedforward operation per sample. AFR adds an extra cost of \(O(N\,\xi_{fw})\). Evidential Alignment also efficiently uses last-layer training in both stages. First, the second-order risk minimization stage runs for \(T_1\) epochs. Then, last-layer retraining runs for \(T_2\) epochs. The overall complexity is \(O(N\,T_1\,\xi_{\text{last}} + N\,T_2\,\xi_{\text{last}})\). Although Evidential Alignment has a slightly higher cost than DFR, it shares the same big-O notation as DFR and AFR, which offers a favorable trade-off between computational efficiency and improved group robustness.

\begin{figure}[h]
    \centering
    \includegraphics[width=\linewidth]{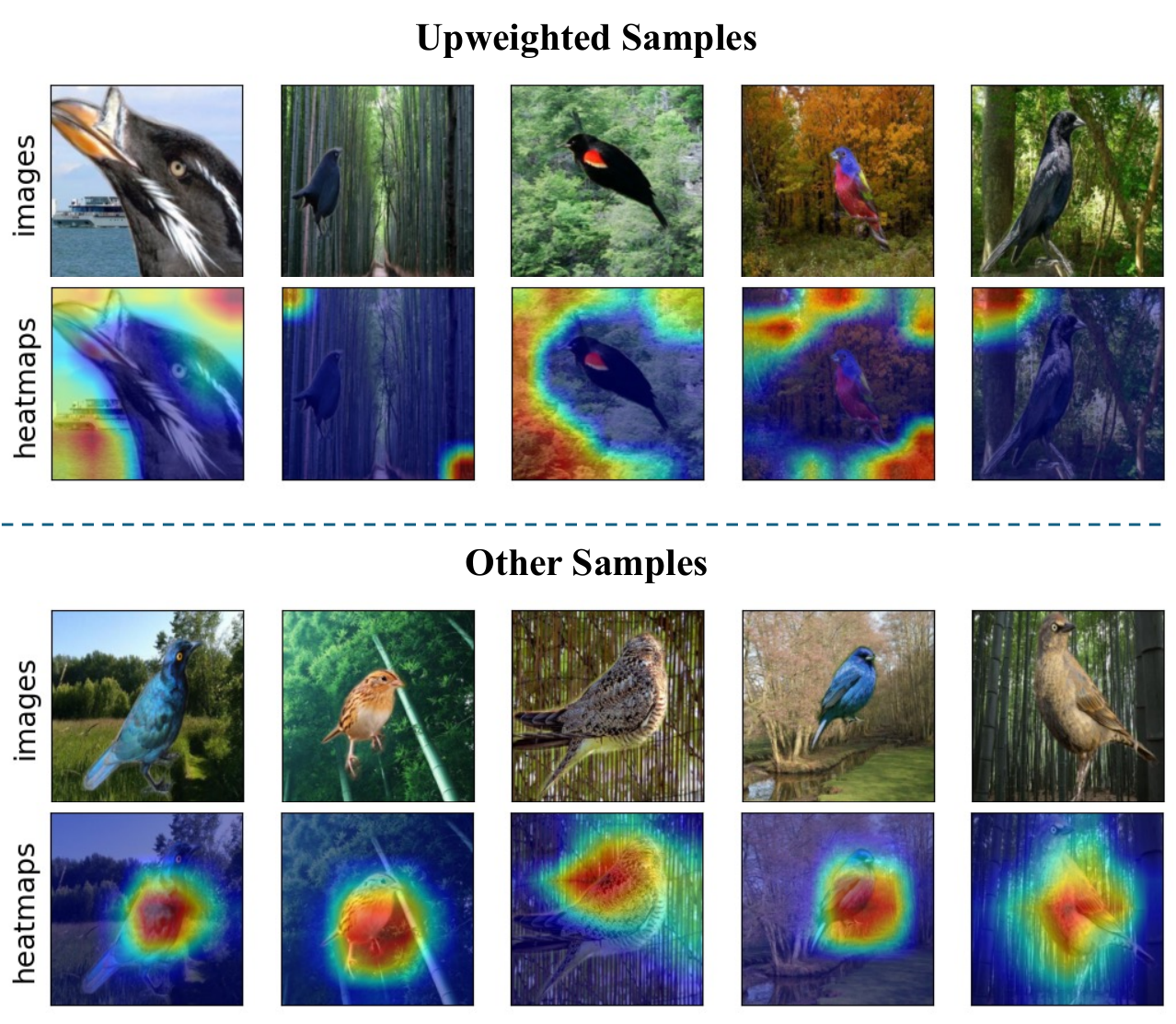}
    \caption{GradCAM~\cite{selvaraju2020grad} visualizations for the upweighted samples and other samples in the Waterbirds dataset.}
    \label{fig:gradcam}
\end{figure}

\subsection{Visualization of Reweighted Samples}
A crucial aspect of Evidential Alignment is determining whether the model’s epistemic uncertainty can effectively reflect the erroneous patterns in the biased ERM model. In our experiments on the Waterbirds dataset, we used GradCAM visualizations (Figure~\ref{fig:gradcam}) to examine the areas of focus for samples with the highest and lowest uncertainty. We observed that high upweighted samples tended to focus background regions associated with the spurious attributes, while non-upweighted samples focused more on the birds themselves, showing our method's effectiveness on identifying potential spurious biases. Additionally, t-SNE embeddings (Figure~\ref{fig:TSNE}) in our synthetic experiment also revealed that high-uncertainty samples clustered around minority groups in the feature space, further supporting the correlation between uncertainty and group information. Quantitative analysis showed correlations between uncertainty values and true group labels across all datasets, indicating that uncertainty estimates are reliable proxies for identifying underrepresented groups. These findings validate our approach of using uncertainty-aware reweighting during retraining, as it allows the model to prioritize learning from challenging and minority group samples, thereby enhancing overall group robustness.

\section{Conclusion}
In this work, we introduce a novel uncertainty-guided group robustness method, Evidential Alignment. Our framework demonstrates that leveraging existing observations from both the biased ERM model and the calibration dataset enables effective mitigation on spurious biases of the deep neural networks. Our method decompose the uncertainty and utilizes epistemic uncertainty from biased ERM models to inform the debiasing process. Extensive experiments demonstrate the efficacy of our approach, achieving favorable worst-group performance across diverse datasets in both vision and text. We believe this work provides valuable insights into developing more adaptive and scalable methods to improve group robustness in real-world applications.

\section*{Acknowledgements}
This work is supported in part by the US National Science Foundation under grants 2217071, 2213700, 2106913, 2008208. Any opinions, findings, and conclusions or recommendations expressed in this material are those of the
author(s) and do not necessarily reflect the views of the National Science Foundation.

\bibliographystyle{ACM-Reference-Format}
\bibliography{sample-base}

\appendix

\section*{\LARGE Appendix}
\section{Dataset Distributions}\label{appendix:data}
By examining the distributions of the four datasets, we can observe distinct patterns of group/class imbalance shown in Table \ref{tab:dataset_quantities_clean}. For the Waterbirds and MultiNLI datasets, although group imbalances are present, the overall class distribution is nearly balanced. In contrast, both group and class imbalances are observed in the CelebA and CivilComments datasets. The CheXpert dataset differs by including more spurious attributes (combination of race (White, Black, Other) and gender (Male, Female)), reflecting a more complex real-world scenario.  This variation suggests that applying simple techniques, such as class-balancing or group-balancing, may have different impacts on improving group robustness, depending on the specific characteristics of the dataset. Recent work by \citet{labonte2024group} also emphasizes that existing class-balancing strategies can produce inconsistent results across these datasets. This underscores the need for a deeper understanding of group robustness, particularly in real-world applications. Consequently, it is crucial to develop adaptive methods that can more effectively address both class and group imbalances to enhance group robustness across diverse datasets.
 
\begin{table}[h]
    \caption{Data quantities and group compositions for each dataset, including training, validation, and test splits.}
    \label{tab:dataset_quantities_clean}
    \centering
    \resizebox{\linewidth}{!}{
    \begin{tabular}{lccccl}
    \toprule
    \textbf{Dataset} & \textbf{Class $y$} & \textbf{Spurious $a$} & \textbf{Train} & \textbf{Validation} & \textbf{Test} \\
    \midrule
    \grayrow \multicolumn{6}{c}{\centering\textit{Waterbirds}} \\
    Waterbirds & landbird   & land       & 3498  & 467  & 2225 \\
    Waterbirds & landbird   & water      & 184   & 466  & 2225 \\
    Waterbirds & waterbird  & land       & 56    & 133  & 642  \\
    Waterbirds & waterbird  & water      & 1057  & 133  & 642  \\
    \midrule
    \grayrow \multicolumn{6}{c}{\centering\textit{CelebA}} \\
    CelebA & non-blond  & female     & 71629 & 8535 & 9767 \\
    CelebA & non-blond  & male       & 66874 & 8276 & 7535 \\
    CelebA & blond      & female     & 22880 & 2874 & 2480 \\
    CelebA & blond      & male       & 1387  & 182  & 180  \\
    \midrule
    \grayrow \multicolumn{6}{c}{\centering\textit{MultiNLI}} \\
    MultiNLI & contradiction & no negation & 57498  & 22814 & 34597 \\
    MultiNLI & contradiction & negation    & 11158  & 4634  & 6655  \\
    MultiNLI & entailment    & no negation & 67376  & 26949 & 40496 \\
    MultiNLI & entailment    & negation    & 1521   & 613   & 886   \\
    MultiNLI & neither       & no negation & 66630  & 26655 & 39930 \\
    MultiNLI & neither       & negation    & 1992   & 797   & 1148  \\
    \midrule
    \grayrow \multicolumn{6}{c}{\centering\textit{CivilComments}} \\
    CivilComments & neutral     & no identity & 148186 & 25159 & 74780 \\
    CivilComments & neutral     & identity    & 90337  & 14966 & 43778 \\
    CivilComments & toxic       & no identity & 12731  & 2111  & 6455  \\
    CivilComments & toxic       & identity    & 17784  & 2944  & 8769  \\
    \midrule
    \grayrow \multicolumn{6}{c}{\centering\textit{CheXpert}} \\
    CheXpert & finding    & (race, gender) & 129434 & 185 & 614 \\
    CheXpert & no-finding & (race, gender) & 61593  & 49  & 54  \\
    \bottomrule
    \end{tabular}
    }
\end{table}

\section{More Experimental Results}
\label{appendix:more_expr}
\noindent \textbf{Hyperparameter Selection.} We include the hyperparameter selection for both second-order risk minimization and evidential calibration in Table \ref{tab:more_exp_1} and Table \ref{tab:more_exp_2}, respectively. We select hyperparameters based on the convergence in training and the performance on the validation data. For example, a too small or too large learning rate hinders the convergence of our algorithm. The annealing step $\eta$ and regularization strength $\beta$ are chosen similarly. The overall performance is robust to small variations in these values.

\noindent \textbf{Backbone Models.} In the main paper, we choose ResNet-50 and BERT (base) as the backbones for experiments, as they were commonly used in previous works. To further validate the effectiveness, we added more experiments with ViT-base pretrained on both ImageNet 1K and ImageNet 21K, shown in Table \ref{tab:more_backbones}. The results show that our method is also effective in improving group robustness in a backbone agnostic manner.

\begin{table}[ht]
\caption{Experiments on Waterbirds with more backbones.}
\label{tab:more_backbones}
\begin{tabular}{lllc}
    \toprule
    Backbone & Pretraining & Method & WGA \\
    \midrule
    ResNet-50 & ImageNet 1K & ERM & 72.6 \\
    \grayrow ResNet-50 & ImageNet 1K & Ours & 92.2 \\
    ViT-base & ImageNet 1K & ERM & 49.4 \\
    \grayrow ViT-base & ImageNet 1K & Ours & 86.4 \\
    ViT-base & ImageNet 21K & ERM & 69.9 \\
    \grayrow ViT-base & ImageNet 21K & Ours & 86.8 \\
    \bottomrule
\end{tabular}
\end{table}

\noindent \textbf{Sensitivity analysis.} To ensure a fair comparison with other baselines, we used half of the validation set as the calibration set in the paper. However, our method is flexible and can work with different calibration set sizes. Table \ref{tab:sensitivity} presents experiments with various portions of the validation set as calibration sets on the Waterbirds dataset. The results indicate that a larger and more diverse calibration set further improves group robustness against spurious correlations.

\begin{table}[ht]
\caption{Sensitivity Analysis on the size of calibration sets. Portion denotes the portion of the validation set used for calibration sets.}
\label{tab:sensitivity}
\begin{tabular}{lccc}
\toprule
Portion & Acc. & WGA & Gap \\
\midrule
0.1 & 95.7 & 83.0 & 12.7 \\
0.2 & 95.0 & 89.7 & 5.3 \\
0.3 & 96.3 & 91.1 & 5.2 \\
0.4 & 96.2 & 91.9 & 4.3 \\
0.5 & 95.1 & 92.2 & 2.9 \\
\bottomrule
\end{tabular}
\end{table}

\begin{table*}[ht]
    \caption{Hyperparameters for second-order risk minimization.}
    \label{tab:more_exp_1}
    \centering
    \begin{tabular}{lccccc}
    \toprule
        Hyperparameters & Waterbirds & CelebA & MultiNLI & CivilComments & CheXpert \\ \midrule
        initial learning rate & 3e-3 & 3e-3 & 1e-3 & 1e-3 & 1e-3\\
        epochs & 50 & 100 & 100 & 100 & 100\\
        annealing step ($\eta$) & 10 & 1 & 1 & 1 & 1 \\
        learning rate scheduler & CosineAnnealing & CosineAnnealing & Linear & Linear & Linear\\
        optimizer & SGD & SGD & AdamW & AdamW & SGD \\ 
        weight decay & 1e-4 & 1e-4 & 1e-4 & 1e-4 & 1e-4\\
        batch size & 32 & 128 & 128 & 128 & 128\\
        backbone & ResNet-50 & ResNet-50 & BERT & BERT & ResNet-50 \\ \bottomrule
    \end{tabular}
\end{table*}

\begin{table*}[ht]
    \caption{Hyperparameters for evidential calibration.}
    \label{tab:more_exp_2}
    \centering
    \begin{tabular}{lccccc}
    \toprule
        Hyperparameters & Waterbirds & CelebA & MultiNLI & CivilComments & CheXpert \\ \midrule
        learning rate & 1e-3 & 2e-2 & 1e-5 & 1e-3 & 1e-2\\
        epochs & 100 & 200 & 200 & 200 & 200\\
        regularization strength ($\beta$) & 10 & 10 & 10 & 10 & 10 \\
        optimizer & SGD & SGD & SGD & SGD & SGD \\
        batch size & 128 & 128 & 128 & 128 & 128 \\ 
        backbone & ResNet-50 & ResNet-50 & BERT & BERT & ResNet-50 \\ \bottomrule
    \end{tabular}
\end{table*}

\section{Proofs of Theoretical Analysis}
\subsection{Proof of Theorem \ref{thm:elbo}} 

\begin{proof}
    We start from the KL divergence between the variational approximation \( p(\boldsymbol{\pi} | x, \theta) \) and the true posterior \( p(\boldsymbol{\pi} | y) \):
    \begin{equation}
        \mathrm{KL}\left[p(\boldsymbol{\pi} | x, \theta) \,\|\, p(\boldsymbol{\pi} | y)\right] = \mathbb{E}_{p(\boldsymbol{\pi} | x, \theta)} \left[\log \frac{p(\boldsymbol{\pi} | x, \theta)}{p(\boldsymbol{\pi} | y)}\right].
    \end{equation}
    
    Using Bayes' rule, we get
    \begin{equation}
        p(\boldsymbol{\pi} | y) = \frac{p(y | \boldsymbol{\pi}) p(\boldsymbol{\pi})}{p(y)},
    \end{equation}
    we rewrite the KL divergence as:
    \begin{equation}
        \mathrm{KL}[p(\boldsymbol{\pi} | x, \theta) \,\|\, p(\boldsymbol{\pi} | y)]
        = \mathbb{E}_{p(\boldsymbol{\pi} | x, \theta)} \left[\log \frac{p(\boldsymbol{\pi} | x, \theta)}{P(y | \boldsymbol{\pi}) p(\boldsymbol{\pi})}\right] + \log p(y).
    \end{equation}
    
    Since KL divergence is non-negative, we obtain the evidence lower bound (ELBO):
    \begin{equation}
        \log p(y) \geq \mathbb{E}_{p(\boldsymbol{\pi} | x, \theta)}[\log P(y | \boldsymbol{\pi})] - \operatorname{KL}[p(\boldsymbol{\pi} | x, \theta) \,\|\, p(\boldsymbol{\pi})].
    \end{equation}
    
    Rearranging, we obtain the ELBO:
    \begin{equation}
        \mathcal{L}_{\mathrm{ELBO}} = -\mathbb{E}_{p(\boldsymbol{\pi} | x, \theta)}[\log P(y | \boldsymbol{\pi})] + \operatorname{KL}[p(\boldsymbol{\pi} | x, \theta) \,\|\, p(\boldsymbol{\pi})].
    \end{equation}

    By parameterizing \( p(\boldsymbol{\pi} | x, \theta) \) as a Dirichlet distribution with parameters \( \boldsymbol{\alpha} \), we use known properties of the KL divergence between Dirichlet distributions and obtain:
    \begin{equation}
        \mathcal{L}_{\mathrm{ELBO}} = \psi (\alpha_y) - \psi(\alpha_0) - \log \frac{B(\boldsymbol{\alpha})}{B(\boldsymbol{\gamma})} + \sum_{k=1}^K(\alpha_k-\gamma_k)(\psi(\alpha_k)-\psi(\alpha_0)),
    \end{equation}
    which is the evidence regularization term controlling confidence in predictions and \(\psi(z) = \frac{d}{dz} \ln \Gamma(z)\) is the digamma function.
    Therefore, minimizing the second-order risk optimizes the ELBO, leading to a model that learns evidence from predictions.
\end{proof}


\subsection{Proof of Theorem \ref{thm:pacbayes-worstgroup}}

\begin{proof}
Fix any group $g\in\{1,\dots,|\mathcal{G}|\}$. Since we have $n_g$ i.i.d.\ training samples from group $g$, a standard PAC-Bayes argument for $[0,1]$-bounded loss implies that for any $0<\delta_g<1$, with probability at least $1-\delta_g$ over the sample of group $g$,
\begin{align}
\mathbb{E}_{f \sim Q}[R_g(f)]
\le
\mathbb{E}_{f \sim Q}[\hat{R}_g(f)]
+
\sqrt{
  \frac{
    \mathrm{KL}(Q \| P)
    +
    \ln \bigl(\tfrac{1}{\delta_g}\bigr)
  }{
    2\,n_g
  }
}.
\end{align}
By choosing $\delta_g = \delta / |\mathcal{G}|$ and applying a union bound, we have that, with probability at least $1-\delta$, every group satisfies
\begin{align}
\mathbb{E}_{f \sim Q}[R_g(f)]
\le
\mathbb{E}_{f \sim Q}[\hat{R}_g(f)]
+
\sqrt{
  \frac{
    \mathrm{KL}(Q \| P)
    +
    \ln \bigl(\tfrac{|\mathcal{G}|}{\delta}\bigr)
  }{
    2\,n_g
  }
}
\end{align}
simultaneously. Taking the maximum over $g$ and noting that $n_g \ge n_{\min}$, we obtain
\begin{align}
    &\max_{1 \le g \le |\mathcal{G}|}
\mathbb{E}_{f \sim Q}[R_g(f)] \\
&\le 
\max_{g}
\left\{
  \mathbb{E}_{f \sim Q}[\hat{R}_g(f)]
  +
  \sqrt{
    \frac{
      \mathrm{KL}(Q \| P)
      +
      \ln \bigl(\tfrac{|\mathcal{G}|}{\delta}\bigr)
    }{
      2\,n_g
    }
  }
\right\}\\
&\le 
\max_{g}
\mathbb{E}_{f \sim Q}[\hat{R}_g(f)]
+
\sqrt{
  \frac{
    \mathrm{KL}(Q \| P)
    +
    \ln \bigl(\tfrac{|\mathcal{G}|}{\delta}\bigr)
  }{
    2\,n_{\min}
  }
}.
\end{align}

Since $\hat{R}_w(f)$ is a convex combination of the group-wise empirical risks, let $a_g = \mathbb{E}_{f \sim Q}[\hat{R}_g(f)]$. We then have
\[
\max_g\,a_g
\le
\frac{1}{\min_{g'} w_{g'}} 
\sum_{g=1}^{|\mathcal{G}|} w_g a_g
=
\frac{1}{\alpha}
\mathbb{E}_{f \sim Q}[\hat{R}_w(f)].
\]
Combining these bounds completes the proof:
\[
\max_{1 \le g \le |\mathcal{G}|}
\mathbb{E}_{f \sim Q}[R_g(f)]
\le
\frac{1}{\alpha}
\mathbb{E}_{f \sim Q}[\hat{R}_w(f)]
+
\sqrt{
  \frac{
    \mathrm{KL}(Q \| P)
    +
    \ln \bigl(\tfrac{|\mathcal{G}|}{\delta}\bigr)
  }{
    2\,n_{\min}
  }
}.
\]
\end{proof}

\end{document}